\documentclass[12pt]{article}

\usepackage{amsmath}
\usepackage{amssymb}
\usepackage{xcolor}
\usepackage{graphicx}
\usepackage{authblk}
\usepackage{hyperref}

\begin{document}

\title{SUS backprop: linear backpropagation algorithm for long inputs in transformers}

\author[1]{Sergey Pankov} 
\author[2]{Georges Harik}
\affil[1]{Harik Shazeer Labs}
\affil[2]{Notbad AI Inc}

\maketitle

\begin{abstract}
It is straightforward to design an unbiased gradient estimator that stochastically cuts the backpropagation flow through any part of a computational graph. By cutting the parts that have little effect on the computation, one can potentially save a significant amount of backpropagation computation in exchange for a minimal increase in the stochastic gradient variance, in some situations. Such a situation occurs in the attention mechanism of the transformer architecture. For long sequences, attention becomes the limiting factor, as its compute requirements increase quadratically with sequence length $n$. At the same time, most attention weights become very small, as most attention heads tend to connect a given token with only a small fraction of other tokens in the sequence. These weights become promising targets for cutting backpropagation. We propose a simple probabilistic rule controlled by a single parameter $c$ that cuts back-propagation through most attention weights, leaving at most $c$ interactions per token per attention head. This brings a factor of $c/n$ reduction in the compute required for the attention backpropagation, turning it from quadratic $O(n^2)$ to linear complexity $O(nc)$. We have empirically verified that, for a typical transformer model, cutting about $99\%$ of the attention gradient flow (i.e. choosing $c \sim 25-30$) results in relative gradient variance increase of only about $1\%$ for $n \sim 2000$, and it decreases with $n$. This approach is amenable to efficient sparse matrix implementation, thus being promising for making the cost of a backward pass negligible relative to the cost of a forward pass when training a transformer model on long sequences.

\end{abstract}


\newcommand{\T}{\mathsf T}

\tableofcontents

\section{Introduction}

The transformer architecture \cite{vaswani2017attention} plays a central role in modern machine learning, achieving state of the art results in many natural language processing tasks \cite{patwardhan2023transformers} and beyond \cite{islam2024comprehensive}, finding its applications in computer vision \cite{dosovitskiy2020image}, speech recognition \cite{radford2023robust}, and time-series analysis \cite{zhou2021informer}, among other fields. The success of transformers is attributed to their ability to process sequential data in parallel, and to the attention mechanism which selects relevant context for every part of the input sequence. 

However, a key challenge with the attention mechanism is its computational complexity, which scales quadratically with the sequence length $n$, making it a significant bottleneck for long sequences. This quadratic complexity arises from the need to compute interactions between every pair of tokens in the sequence.

There have been numerous attempts to alleviate this issue, often by modifying the attention mechanism to achieve sub-quadratic complexity. To name just a few: sparse attention based on locality sensitive hashing \cite{kitaev2020reformer} with ${\cal O}(n\log{n})$ complexity, attention with kernel functions in place of standard softmax \cite{katharopoulos2020transformers, choromanski2020rethinking}, and attention based on low-rank approximation \cite{wang2020linformer}, both with linear  ${\cal O}(n)$ complexity. While those implementations demonstrated impressive speedups, the gains were potentially coming with a performance cost, as the originally proposed attention mechanism remains the dominant option for training language models, despite its quadratic cost.

Other approaches, notably Flash Attention \cite{dao2022flashattention,dao2023flashattention,shah2024flashattention}, focused on improving the attention efficiency without altering the underlying algorithm. By moving computations to a smaller but faster part of GPU memory, significant speedups were gained for both forward and backward model passes. 

Likewise, the method that we propose does not change the attention mechanism. Our goal is to alleviate the computational bottleneck in attention's backpropagation by changing it from quadratic to linear for long sequences. This is achieved by stochastically introducing sparsity in the gradient flow through attention, without introducing a bias in gradient estimation. Accordingly, it is called: Sparse Unbiased Stochastic backpropagation (SUS backprop). The paper provides empirical evidence that SUS backprop achieves a substantial reduction in computational cost with minimal impact on gradient variance. 

The paper is organized as follows. Section \ref{susbackprop} introduces the approach and related formalism. Section \ref{results} provides empirical insights in attention's behavior in transformers and demonstrates a favorable sparsity-variance tradeoff of SUS backprop. Related work is discussed in Section \ref{related} with a future work outlook given in Section \ref{discussion}.

\section{SUS backprop: Sparse Unbiased Stochastic backprop}
\label{susbackprop}

Consider a computation graph $G$ (which is necessarily a directed acyclic graph), where each node $i$ is a function $f_i(f_j,...)$ that takes as its arguments the values $f_j,...$ of its parent nodes that have been already computed in accordance with the topological ordering of $G$. Let $ji$ denote an edge directed from $j$ to $i$. Let $p$ denote a path in $G$ (traversed in the direction of its edges). Let $\bar{ij}$ and $\bar p$ be the reversed edge $ji$ and reversed path $p$, respectively. Let us assign to every edge $ji$ the partial derivative $\partial f_i / \partial f_j$ of the node $i$ with respect to its parent $j$, (one can think of $\partial f_i / \partial f_j$ as a Jacobian matrix, in case of vector-valued inputs and outputs of $f$). Then the full derivative of any node $k$ with respect to any other node $l$ can be computed as a sum of products of partial derivatives along all the possible paths $P_{lk}$ in $G$ connecting $l$ to $k$:
\begin{equation}
  \frac{d f_k}{d f_l} = 
  \sum_{p\in P_{lk}} \prod_{\bar{ij} \in \bar{p}} \frac{\partial f_i}{\partial f_j}.
  \label{dfdf}
\end{equation}
Since no edge can be traversed twice by a path in $G$, it follows that the full derivative is a linear function of any partial derivative $\partial f_i / \partial f_j$. Therefore, the expression in Eq.(\ref{dfdf}) can be stochastized by multiplying every partial derivative $\partial f_i / \partial f_j$ by a random variable $\tilde m_{ij}$, which plays the role of an upweighting stochastic mask, defined as
\begin{equation}
  \tilde m = \quad 
  \left\{
  \begin{split}
    & \frac{1}{q} \, \text{, with probability } q, \\
    & 0 \, \text{, with probability } 1-q,
  \end{split}
  \right.
  \label{tildeq}
\end{equation}
without affecting the expectation value 
\begin{equation}
  \mathbb{E}\left[
  \sum_{p\in P_{lk}} \prod_{\bar{ij} \in \bar{p}} 
  \frac{\partial f_i}{\partial f_j} \tilde m_{ij}
  \right]
  = \frac{d f_k}{d f_l}.
  \label{stoch_dfdf}
\end{equation}
Moreover, the same random variable can be used with different partial derivatives, as long as they cannot be connected by a path, (otherwise, the same random variable will occur more than once on the path, thus entering non-linearly the full derivative and invalidating Eq.(\ref{stoch_dfdf})). More generally, any coupling of random variables that is linear (in each random variable), with the random variables satisfying Eq.(\ref{tildeq}), will result in an unbiased stochastic estimator of full derivatives. For example, the coupling could be done internally inside $\partial f_i / \partial f_j$, as long as $\partial f_i / \partial f_j$ remains linear in random variables. 

Thanks to the unbiased gradient estimation, we expect our approach to improve computational complexity without fundamentally altering the optimization trajectory during training, in that sense being equivalent to the original attention mechanism. Also, the unbiased gradient generally offers stronger convergence guarantees \cite{vicol2021unbiased}.

\subsection{SUS backprop through attention}

We now consider a concrete example of the transformer attention mechanism. Let $n$ be the sequence length and $d$ be the attention head dimension. For a single attention head, the attention block takes queries $Q$, keys $K$ and values $V$ as inputs (all three are $\in \mathbb{R}^{n \times d}$), and outputs a weighted average of values $\bar V$:
\begin{equation}
\bar V = W V,
\end{equation}
where the attention weights $W \in \mathbb{R}^{n \times n}$ are
\begin{equation}
  W = \text{softmax}\left(QK^\T\right).
  \label{W}
\end{equation}
The softmax is applied row-wise, so $\sum_j W_{ij} = 1$. Let the gradient of $a$ with respect to $b$ be $\nabla_b a \equiv  da / db$. Let $\cal L$ be the model loss. For the gradient of the loss with respect to the attention inputs we find (see Appendix \ref{attngrads} for details):
\begin{equation}
  \begin{split}
    & \nabla_Q {\cal L} = M K, \\
    & \nabla_K {\cal L} = M^\T Q, \\
    & \nabla_V {\cal L} = W^\T  \nabla_{\bar V} {\cal L},
  \end{split}
  \label{loss_grad}
\end{equation}
where we have defined the matrix $M$ as
\begin{equation}
  M_{ij} = W_{ij}\sum_{\nu} \left( V_{j\nu} - \bar V_{i\nu} \right)
  \nabla_{\bar V_{i\nu}} {\cal L}.
  \label{M}
\end{equation}
We stochastize $\nabla {\cal L}$ by stochastizing $W \to \tilde W = W \odot \tilde m$ where it explicitly enters Eqs.(\ref{loss_grad},\ref{M}). (Note that all the forward-computed quantities, e.g. $\bar V$, are unchanged by this procedure.) There is no restriction on how the acceptance probabilities $q_{ij}$ are computed. We chose a very simple rule, controlled by a single parameter $c$:
\begin{equation}
  q_{ij} = \min\{c W_{ij}, 1\}.
\end{equation}
Because $\sum_j q_{ij} \le c\sum_j W_{ij} = c$, we call $c$ the attention retention parameter, as it plays the role of the upper bound (which is saturated for $c \le 1$) on the expected number of attention interactions retained by the above rule, per token per attention head. Since only the retained weights are counted in the gradient computation in Eqs.(\ref{loss_grad},\ref{M}), and the masking decisions in $\tilde W$ can be made in the forward run, then by controlling the sparsity of $\tilde W$, $c$ also controls the memory and compute complexity of the back-propagation through attention. It takes ${\cal O}(n c)$ memory to store $\tilde W$ and ${\cal O}(n c d)$ operations to compute the gradient. Therefore, both time and space complexities are reduced by factor $c/n$ (assuming that $W$ is passed from the forward run, rather than recomputed). If $W$ is recomputed in the backward run, then we have the same reduction in compute, while paying ${\cal O}(n c)$ in extra memory needed to pass $\tilde W$. If $c$ is small relative to $d$, then this extra memory demand is inconsequential as we already pay ${\cal O}(n d)$ for passing $(Q,K,V)$. One way or the other, the complexity of SUS backprop is linear in $n$, making it superior to any quadratic complexity algorithm for long sequences, provided the benefits of sparsity outweigh the detriments of the gradient variance increase. In this paper, we present compelling evidence in support of that assumption.

\subsection{Toy model of sparsity-variance tradeoff}
\label{toymodel}

To gain a theoretical insight into the sparsity-variance tradeoff, we consider a toy model of the attention contribution to the gradient variance. We call it a $k$-weight model, because it features $k$ distinct attention weight values. The simplest case of $k = 2$, despite its crudeness, already provides a qualitative picture of the tradeoff, as shown in Appendix \ref{kweight}.
 
Let us model the attention contribution from a single token, single head, single layer perspective as
\begin{equation}
  \nabla {\cal L} = \sum_{a=1}^{a=k} w_a\sum_{b=1}^{b=m_a} v_{ab},
  \label{toymodelgrad}
\end{equation}
where $m_a$ is the multiplicity of each attention weight value $w_a$ and $v_{ab}$ are independent variables of unit variance. We assume that $w_a$ are ordered in increasing order, for convenience. Note that the form of the above equation is not limited to a single layer case. Due to the multiplicative nature of probabilities, due to the gradient being a sum of products (see Eq.(\ref{dfdf})) and due to $w$ and $q$ being probabilities, the same toy model can describe a multi-layer network.

For convenience, we work with reduced quantities, where the attention weights, weight multiplicities, attention retention parameter, attention retention and gradient variance are all scaled with $n = \sum_{a=1}^{a=k} m_a$. Specifically: $\omega_a = w_a n$, $\mu_a = m_a/n$, $\xi = c/n$, and $\kappa$ and $\Sigma$ are the reduced attention retention and gradient variance, respectively.

For $k$-weight model, we derived a simple sparsity-variance tradeoff relationship (see Appendix \ref{kweight}), that is conveniently expressed in terms of the slopes of $\kappa(\xi)$ and $\Sigma(\xi)$ and that can be tested on empirical data:
\begin{equation}
  \frac{d\kappa}{d\xi} = -\xi^2 \frac{d\Sigma}{d\xi} .
  \label{kappasigmarelation}
\end{equation}
Let $\Sigma_0$ be the original stochastic gradient variance, which is controlled by the minibatch sampling choices and coincides with $\Sigma$ in the limit $\xi \to \infty$. In case of a power-law dependence of $\kappa \propto \xi^\alpha$ and $\Sigma - \Sigma_0 \propto \xi^\beta$, the above equation implies $\alpha - \beta = 2$. How well this relationship is respected by actual language model data is tested in Section \ref{empirical}.

\section{Experiments and results}
\label{results}

Our idea, that backpropagation can be cut through most attention weights without causing a significant increase in the gradient variance, was motivated by an assumption that a typical token interacts via the attention mechanism with a relatively small number of other tokens \cite{bahdanau2014neural,clark2019does}, rendering most attention weights small. To gain understanding, to what extent this assumption is correct \cite{araabi2024entropy} or not \cite{zhai2023stabilizing}, we will take a closer look at the distribution of the attention weights across the heads, layers, language models and input sequence lengths \cite{clark2019does}, in the first subsection of this section. In the second subsection, we directly measure the dependence of the gradient variance increase on $c$ and $n$ in a language model with a custom implementation of SUS backprop algorithm. We will also compare the results with $k$-weight model predictions.

\subsection{Attention spread in transformer models}

Let us define quantities that help us characterize how much the distribution of attention weights is spread or peaked. Let top-$\mathsf p$ weights be the top weights, whose combined probability mass is $\mathsf p \in [0, 1]$, (the threshold parameter is fixed throughout this section to $\mathsf p = 0.9$). Let the attention spread $s_i$ for a token at position $i$ be the number of top-$\mathsf p$ weights among $W_{ij}$. The spread fraction $s_i/i$ indicates the fraction of tokens that significantly contribute to the attention output $\bar V_i$, (contribute on average, assuming statistical independence of $V_j$). Let us define the aggregate spread fraction $\phi_i$ as:
\begin{equation}
\phi_i = \frac{\sum_{j=0}^{j=i} s_j}{\sum_{j=0}^{j=i} j}.
\end{equation}
The spread fraction $\phi_i$ describes the fraction of top-$\mathsf p$ reduced weights $\omega_{jk} = (j+1) W_{jk}$ among all the weights with $k \le j \le i$. While $s_i$ is suitable for describing the attention of the token at position $i$, $\phi_i$ is suitable for describing the attention of all the tokens up to position $i$. As an aggregate quantity, $\phi_i$ is much less sensitive to statistical noise, than $s_i$. 

We measure $s_i$ and $\phi_i$ by averaging over 1000 input sequences. Four models -- {\it opt-125m, opt-350m, opt-1.3b,} \cite{zhang2022opt} and {\it Mistral-7B-v0.1} \cite{jiang2023mistral7b} -- were tested on the sequences from {\it The Pile} dataset \cite{gao2020pile}. The first three smaller models were tested on the full span of permitted input length (2048 tokens), while Mistral could be tested up to maximum 1570 tokens (on A100 GPU).

\begin{figure}[h]
  \centering
  \begin{minipage}{0.49\textwidth}
    \centering
    \includegraphics[width=\textwidth]{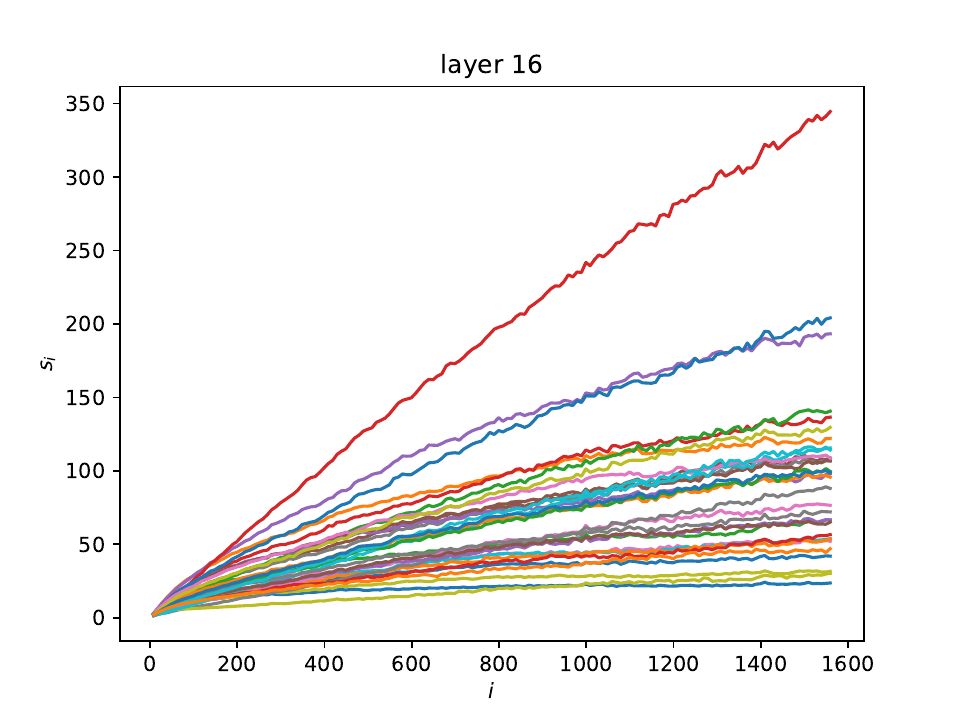}
    \caption{Attention spread $s_i$ vs token position $i$ for 32 attention heads of layer 16, Mistral model.}
    \label{layer16mistral}
  \end{minipage}
  \hfill
  \begin{minipage}{0.49\textwidth}
    \centering
    \includegraphics[width=\textwidth]{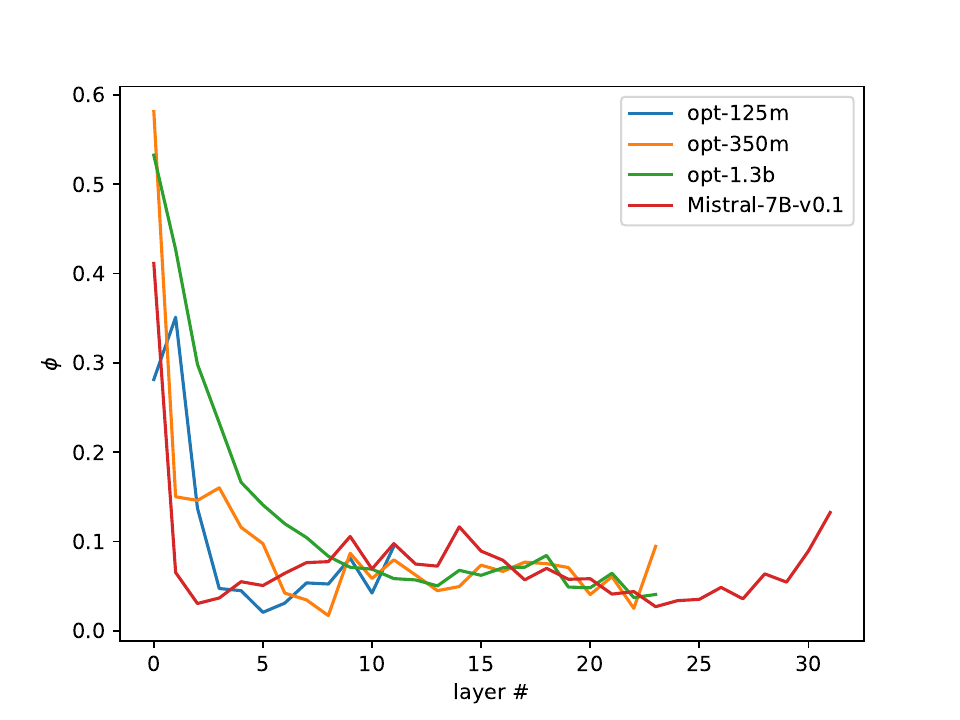}
    \caption{Aggregate spread fraction $\phi_i$, averaged over layer's heads, vs layer index for different models. For OPT models, $i = 2000$; for Mistral, $i = 1550$.}
    \label{spreadvslayers}
  \end{minipage}
\end{figure}

\begin{figure}[h]
  \centering
  \begin{minipage}{0.49\textwidth}
    \centering
    \includegraphics[width=\textwidth]{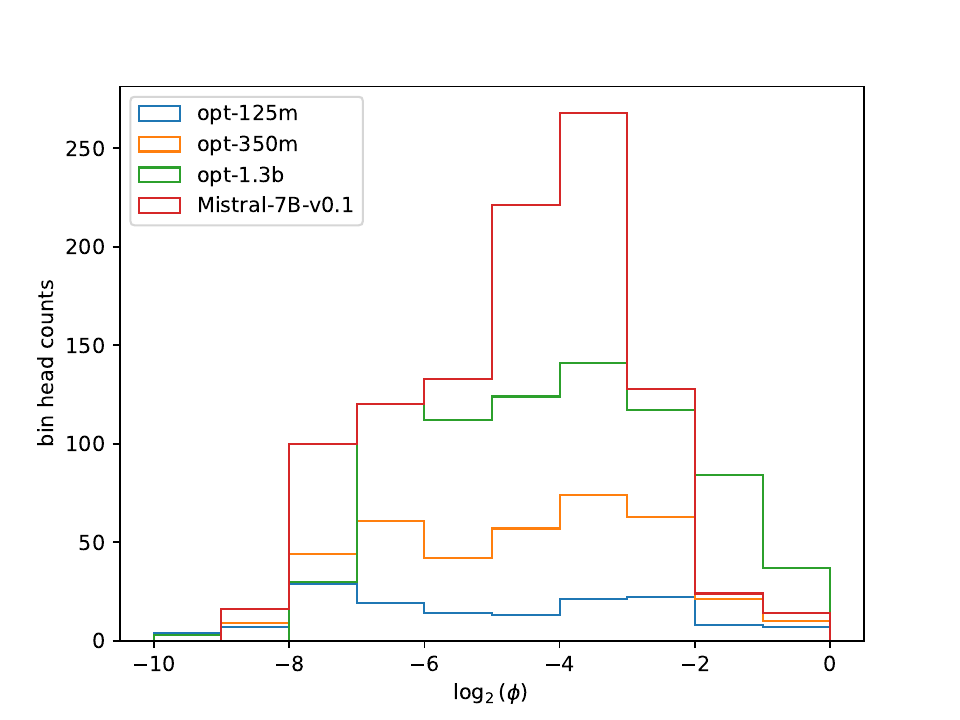}
    \caption{Distribution of attention head log-spreads $\log_2{\phi_i}$ in different models. For OPT models, $i = 2000$; for Mistral, $i = 1550$.}
    \label{distrheadspread}
  \end{minipage}
  \hfill
  \begin{minipage}{0.49\textwidth}
    \centering
    \includegraphics[width=\textwidth]{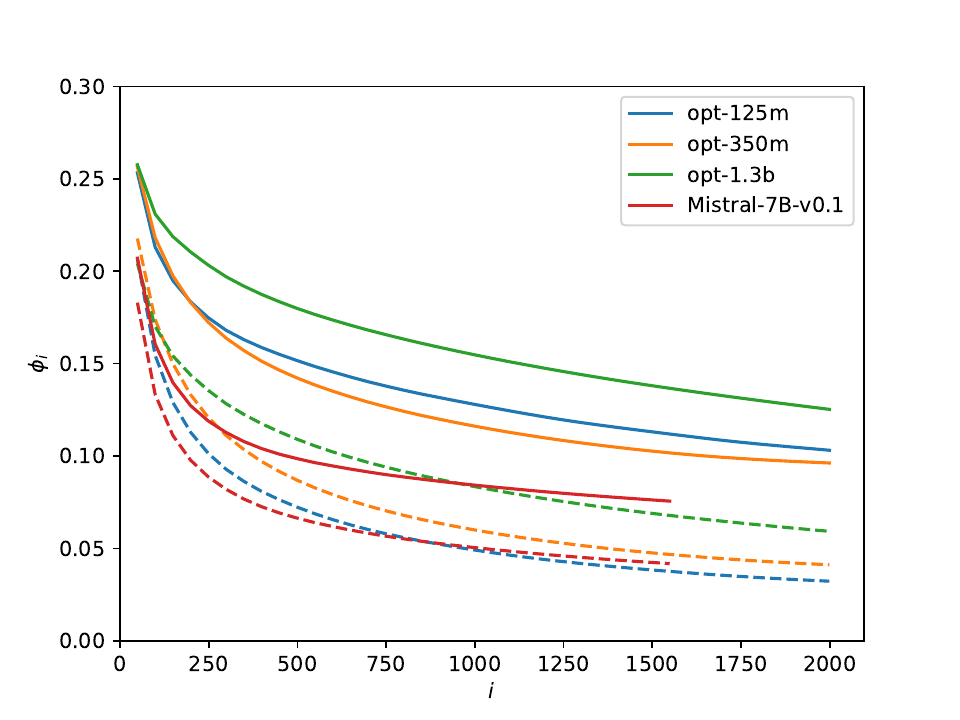}
    \caption{Aggregate spread fraction $\phi_i$, averaged over all heads and layers, vs token position $i$ for different models. Solid and dashed lines of the same color represent arithmetic and geometric mean, respectively, for the same model.}
    \label{spreadvslength}
  \end{minipage}
\end{figure}

Measuring the attention spread $s_i$ across different attention heads of different layers of different models, we observed a great variation in the spread behavior. While it is impossible to reflect all that diversity in a single illustration, a somewhat common situation in a layer is shown in Fig.(\ref{layer16mistral}), (see Appendix \ref{attnspreadmodels} for a complete account of all layers in all models). Typically, the majority of heads show a limited spread $s_i$, which increases sub-linearly with $i$. Still, in a few heads the attention is spread much wider, with the spread trending closer to linear. The distribution of heads by their spread is shown in Fig.(\ref{distrheadspread}), (note the logarithmic scale of the spread). The heads with wider spread are less numerous, but they disproportionately contribute to the arithmetic mean of the spread, while the more focused heads, which are more numerous, dominate the geometric mean of the spread. These two measures of attention spread are noticeably different, with the geometric mean being significantly lower, see Fig.(\ref{spreadvslength}). Notice how the ratio of the arithmetic to geometric means keeps increasing  with the sequence length. Encouragingly, the largest and most advanced model demonstrates overall the lowest attention spread.

The presence of heads with a large attention spread may potentially seem problematic for SUS backprop, since, at face value, they violate the assumption of every token talking to only a few counterparts. In practice, cutting backpropagation through the majority of their seemingly equally important weights did not appear to cause much problem, as follows from the results of the next section. We can only speculate on the reason underlying that insensitivity: either information coming from various values $V_i$ is greatly redundant for those heads, or they simply do not extract valuable information \cite{voita2019analyzing}.

The evolution of aggregate attention spread fraction $\phi$ across model's layers reveals a loosely similar pattern \cite{clark2019does} in different models. The spread is largest at the lowest layers, then it quickly drops with the layer index increasing, remaining approximately  at its lowest values (about $\sim 0.1$ for the mean spread for $n \sim 2\times 10^3$) through most of the layers, finally showing a slight increase at the topmost layers, see Fig.(\ref{spreadvslayers}).

Additionally, we observed various artifacts, such as intricate, non-monotonic behavior of some heads, especially in the OPT series, see Appendix \ref{attnspreadmodels}. In some heads the spread saturates, in some - starts decreasing after an initial increase, yet in some other heads it follows roughly oscillatory patterns of various periodicities. The occurrence of those heads does not follow any obvious pattern, appearing at different layers and model sizes, in the models sharing the same architecture.
 
\subsection{Empirical observation of sparsity-variance tradeoff}
\label{empirical}

We have implemented SUS-backprop algorithm by modifying Hugging Face's attention code for OPT model. The part of the code, that computes the attention weights is replaced by a function with a custom backward() method, that implements the loss gradient with respect to $(Q, K, V)$ as described in Eqs.(\ref{loss_grad},\ref{M}). The forward() method implements the weight matrix $W$ stoshastization, saving for the backward pass a sparse matrix $\tilde W$, requiring ${\cal O}(nc)$ memory (in sparse representation). For testing purposes, the algorithm is implemented in both dense and sparse representations.

\begin{figure}[h]
  \centering
  \begin{minipage}{0.49\textwidth}
    \centering
    \includegraphics[width=\textwidth]{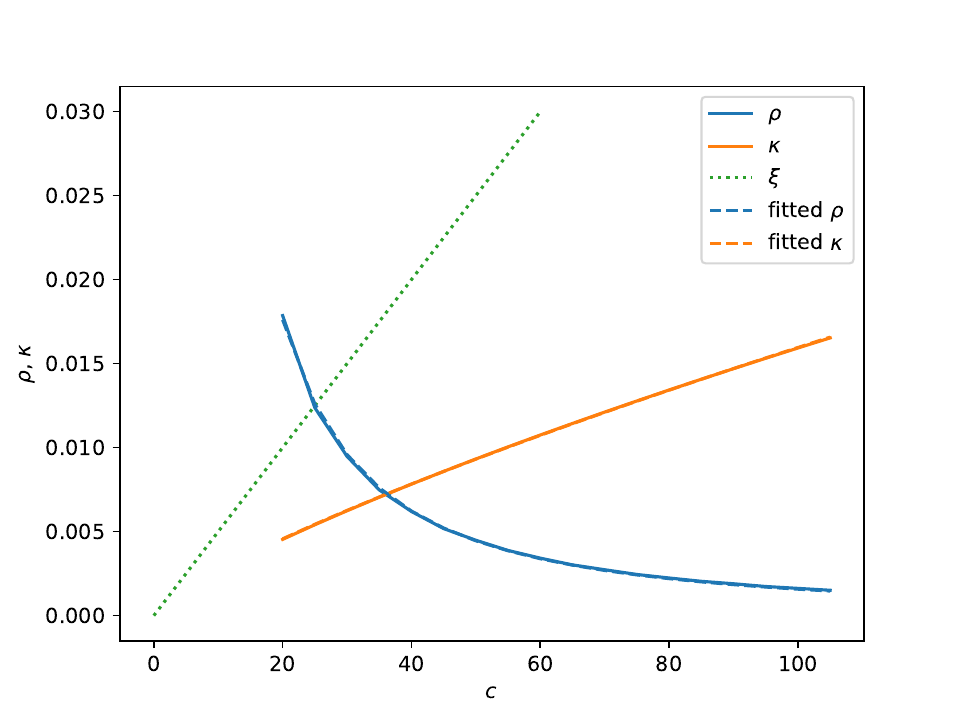}
  \end{minipage}
  \hfill
  \begin{minipage}{0.49\textwidth}
    \centering
    \includegraphics[width=\textwidth]{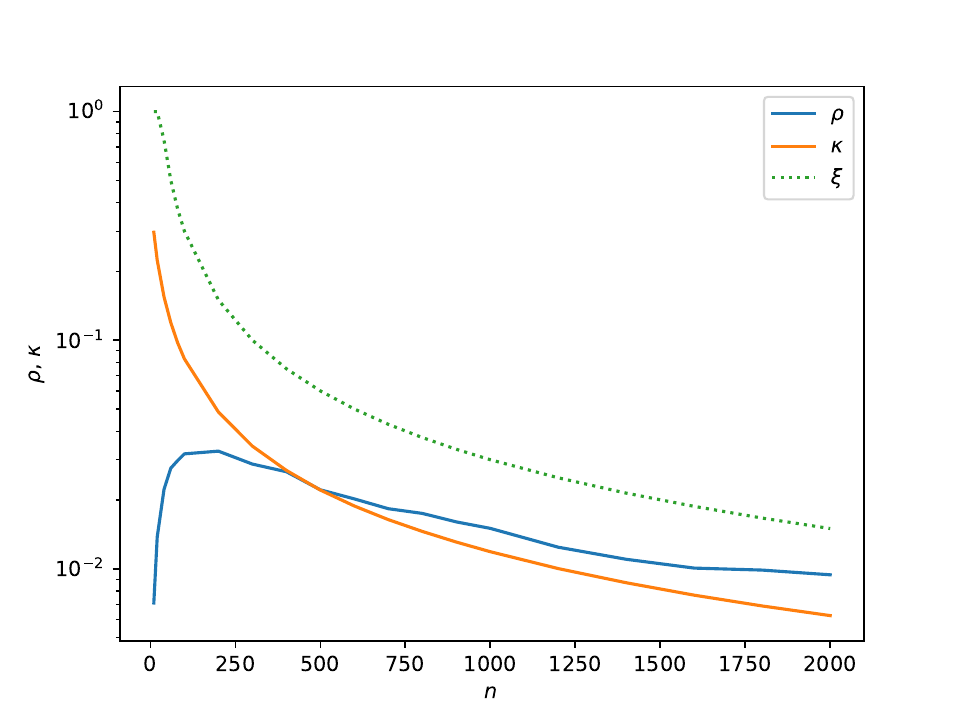}
  \end{minipage}
  \caption{opt-125m: relative gradient variance increase $\rho$ and reduced attention retention $\kappa$. The dotted line $\xi$ is the upper bound on $\kappa$. Left panel: $\rho$ and $\kappa$ vs $c$ for sequence length $n = 2000$. The dashed lines are power-law fits: $\rho \propto \xi^{-1.5}$ and $\kappa \propto \xi^{0.78}$. Right panel: $\rho$ and $\kappa$ vs $n$ for attention retention parameter $c = 30$.}
  \label{rhokappavscn}
\end{figure}

From practical standpoint, the most important results are presented in this section, where we test how much the sparsity of $\tilde W$ affects the relative increase in gradient variance, which we define as
\begin{equation}
  \rho = \frac{\Sigma - \Sigma_0}{\Sigma_0}.
\end{equation}
For our approach to have a practical significance, we ideally would like to see that for $c \lesssim d$ (implying a moderate demand in additional memory for storing $\tilde W$) we have a relatively small $\rho \ll 1$. Indeed, for sufficiently large $n$, a linear time complexity backprop algorithm should give a factor of $2-3$ speedup, by making the backward pass time negligible in comparison to the forward pass time. But if this speedup is accompanied by a noticeable increase in variance, say, $\rho \sim 1$, it may negate the gains in compute per data sample, as we would need to process significantly more data samples, to counter the effects of the increased gradient variance. Fortunately, we found that for $c \sim 25-30$ and $n = 2000$ we already have $\rho \sim 0.01$, and importantly, $\rho$ keeps decreasing with increasing $n$. This is shown in Fig.(\ref{rhokappavscn}).

We measure the gradient variance as a mean variance over all gradient components, averaged over 100 input sequences for every data point in the plots. In the left panel of Fig.(\ref{rhokappavscn}) the variance increase $\rho$ and the attention retention $\kappa$ are plotted against the attention retention parameter $c$ for a fixed $n = 2000$. It also shows rather accurate power-law fits of $\rho \propto \xi^\beta$ and $\kappa \propto \xi^\alpha$, which let us test the relationship $\alpha - \beta = 2$, predicted by $k$-weight model in Section \ref{toymodel}. We found $\alpha \approx 0.78$ and $\beta \approx -1.50$, resulting in $\alpha-\beta = 2.28$, which is not too far off from the theoretical prediction. See more in Appendix \ref{kweight} on $2$-weight model fitting.

In the right panel of Fig.(\ref{rhokappavscn}) $\rho$ and $\kappa$ are plotted against the input sequence length $n$ for a fixed $c = 30$. It shows that $\rho$ initially quickly increases with $n$, reaching the maximum value of about $\sim 0.033$ at $n \sim 200$, but then the trend reverses and $\rho$ keeps decreasing with $n$, reaching just under $0.01$ at $n \sim 2000$. Assuming this trend continues for larger $n$ and is generic for transformer models, the above results provide strong evidence that SUS backprop can bring significant compute savings for training with long sequences.

\section{Related work}
\label{related}

Our formulation of SUS backprop is directly applicable to the original Multi-Head Attention (MHA) mechanism \cite{vaswani2017attention}, which treats the heads as a disjoint set of $(Q, K, V)$ triplets. There have been proposed multiple attention mechanism modifications, that fundamentally do not change its quadratic complexity ${\cal O}(n^2)$, but can realize a more economical use of memory and compute resources. Some of the most notable variations include Multi-Query Attention (MQA) \cite{shazeer2019fast}, Grouped-Query Attention (GQA) \cite{ainslie2023gqa} and Multi-head Latent Attention (MLA) \cite{liu2024deepseek}. MQA, prominently used in PaLM model \cite{chowdhery2023palm}, employs multiple queries $Q$, but shared $KV$ pairs. GQA (used by LLaMA \cite{touvron2023llama} and Mistral \cite{jiang2023mistral7b}, among others) interpolates between MHA and MQA, by mapping groups of queries to distinct $KV$ pairs. MLA (used by DeepSeek \cite{liu2024deepseek}) resorts to low rank compression of $KV$ matrices, that reduces memory footprint. Our method is in principle compatible with these attention variants, but would require appropriate modification of attention gradient equations Eqs.(\ref{loss_grad},\ref{M}).

A recently proposed Native Sparse Attention \cite{yuan2025native}, offering an order of magnitude speedups over standard full attentions (such as MQA and GQA) thanks to the use of compressed token blocks and hardware-efficient implementation, is still asymptotically a quadratic complexity algorithm, which can potentially benefit from backpropagation sparsification.

There also have been proposed many attention mechanisms with nominal ${\cal O}(n)$ complexity. Some mechanisms are based on approximate search (Reformer \cite{kitaev2020reformer}), others rely on low-rank approximations (Linformer \cite{wang2020linformer}), kernel methods to replace softmax (Performer \cite{choromanski2020rethinking}) or sparsity (Longformer \cite{beltagy2020longformer}, BigBird \cite{zaheer2020big}). These approaches share an idea that explicit computation of negligible (or otherwise unimportant) attention weights can be avoided during the forward run. In which case, the gradient flow in the backward run is unlikely to be sparse, so there would be little room for improvement left for a backpropagation sparsification algorithm, such as SUS backprop. However, the competitiveness of forward ${\cal O}(n)$ algorithms remains to be convincingly demonstrated.

A stochastic sparsification of backpropagation was employed in Refs. \cite{cheng2022stochastic, fang2022depth} in the context of training video models, but unlike in our work, no upweighting was used to ensure unbiased gradient estimation. Also, the gradient flow was cut randomly, relying on redundancy naturally present in video inputs, rather than selecting parts that have little effect on forward computation, as in our work. 

\section{Discussion}
\label{discussion}

In this paper, we presented a novel backpropagation algorithm -- SUS backprop -- that sparsifies the flow of gradient through attention in a stochastic and unbiased manner. We presented a compelling evidence, that this sparsification can transform the computational complexity of the backward pass from quadratic to linear, at the cost of only a small increase in gradient variance.

Yet more work remains to be done. To make this approach practically useful and competitive, it needs a more efficient implementation. Our current implementation fully relies on native PyTorch constructs from its sparse matrix module. This was sufficient to prove the concept and to demonstrate a linear compute complexity of the backward pass over quadratic complexity of the forward pass. However,  due to serious limitations of PyTorch sparse matrix support, we incurred unnecessarily large overhead, offsetting the gains from backpropagation by the slowdown in the forward run. It will most likely require writing a custom kernel to overcome the current sparse module limitations.

While our analysis shows a very favorable sparsity-variance tradeoff, both in sequence length $n$ and sparsity control parameter $c$, indicating a great potential of SUS backprop for long sequence training, the ultimate verdict should come from actual training results, be it a supervised fine tuning or full fledged pretraining of a language model. It may also be interesting to observe the evolution of attention spread during pretraining, by analyzing available checkpoints of large language models.

\bibliographystyle{plain}

\bibliography{attn_refs}

\appendix

\section{Attention gradients}
\label{attngrads}

We want to compute the gradient of the loss function $\cal{L}$ with respect to the attention inputs $Q$, $K$ and $V$. In matrices, we use Latin letters for token position indices and Greek letters for head dimension indices. To simplify notations, we may omit some matrix indices or gradient subscripts, when it does not create a confusion. To compute
\begin{equation}
  \nabla {\cal L} = \sum_{j\nu} \left( \nabla {\bar V_{j\nu}} \right) \nabla_{\bar V_{j\nu}} {\cal L},
  \label{loss_grad_gen}
\end{equation}
we need to compute:
\begin{equation}
  \nabla {\bar V_{j\nu}} = \sum_k \left( \left(\nabla W_{jk}\right) V_{k\nu} 
  + W_{jk} \nabla V_{k\nu} \right).
  \label{attn_grad_gen}
\end{equation}
The attention weights, given by Eq.(\ref{W}), are:
\begin{equation}
  W_{jk} = \frac{e^{S_{jk}}}{\sum_l e^{S_{jl}}}, \quad S_{jk} = Q_j K_k^\T.
  \label{Wjk}
\end{equation}
Note that $\nabla_Q V = \nabla_K V = \nabla_V W = 0$. Differentiating $W$ with respect to $Q$ and $K$ we find:
\begin{equation}
  \begin{split}
    & \nabla_{Q_{i}} W_{jk} = 
    \delta_{ij} W_{jk} \left( K_{k} - \sum_l W_{jl} K_{l} \right), \\
    & \nabla_{K_{i}} W_{jk} = 
    W_{jk} \left( \delta_{ik} - W_{jl} \right) Q_{j}.
  \end{split}
  \label{w_grad}
\end{equation}
Also,
\begin{equation}
  \nabla_{V_{i\mu}} V_{k\mu} = \delta_{ik} \delta_{\mu\nu}.
  \label{v_grad}
\end{equation}
From Eq(\ref{attn_grad_gen}) and Eqs(\ref{w_grad},\ref{v_grad}) we find:
\begin{equation}
  \begin{split}
    & \nabla_{Q_{i\mu}} \bar V_{j\nu} =
    \delta_{ij} \sum_{k} W_{jk} \left( V_{k\nu} - \bar V_{j\nu} \right) K_{k\mu}, \\
    & \nabla_{K_{i\mu}} \bar V_{j\nu} = 
    W_{ji} \left( V_{i\nu} - \bar V_{j\nu} \right) Q_{j\mu}, \\
    & \nabla_{V_{i\mu}} \bar V_{j\nu} = W_{ji} \delta_{\mu\nu}.
  \end{split}
  \label{attn_grad}
\end{equation}
From Eq(\ref{loss_grad_gen}) and Eq(\ref{attn_grad}) we find:
\begin{equation}
  \begin{split}
    & \nabla_{Q_{i\mu}} {\cal L} = 
    \sum_{j} W_{ij} \sum_{\nu} \left( V_{j\nu} - \bar V_{i\nu} \right) K_{j\mu}
    \nabla_{\bar V_{i\nu}} {\cal L}, \\
    & \nabla_{K_{i\mu}} {\cal L} = 
    \sum_{j} W_{ji} \sum_{\nu} \left( V_{i\nu} - \bar V_{j\nu} \right) Q_{j\mu}
    \nabla_{\bar V_{j\nu}} {\cal L}, \\
    & \nabla_{V_{i\mu}} {\cal L} = 
    \sum_{j} W_{ji} \nabla_{\bar V_{j\mu}} {\cal L}.
  \end{split}
\end{equation}
From where the equations Eqs(\ref{loss_grad},\ref{M}) in Section \ref{susbackprop} follow.

\section{$k$-weight model}
\label{kweight}

The weights and their multiplicities in $k$-weight model are constrained by the following equations, expressing that multiplicities sum up to $n$ and weights sum up to 1, respectively:
\begin{equation}
  \begin{split}
    & \sum_{a=1}^{a=k} \mu_a = 1, \\
    & \sum_{a=1}^{a=k} \mu_a \omega_a = 1.
  \end{split} 
\end{equation}
Attention retention $\kappa$ (which is the expectation of the accepted weights) and gradient variance $\Sigma$ (which is the variance of the upweighted accepted terms of the gradient in Eq.(\ref{toymodelgrad})) are expressed in terms of multiplicities $\mu_a$, weights $\omega_a$ and acceptance probabilities $q_a$ as 
\begin{equation}
  \begin{split}
    & \kappa = \sum_{a=1}^{a=k} \mu_a q_a, \\
    & \Sigma = \sum_{a=1}^{a=k} \frac{\mu_a \omega_a^2}{q_a},
  \end{split} 
\end{equation}
where $q_a = \min\{\xi\omega_a, 1\}$. Let $\xi$ be in the interval $1/\omega_{l} \le \xi \le 1/\omega_{l-1}$. Then:
\begin{equation}
  \begin{split}
    & \kappa = \xi \sum_{a=1}^{a<l} \mu_a \omega_a + \sum_{a=l}^{a=k} \mu_a, \\
    & \Sigma = \frac{1}{\xi}\sum_{a=1}^{a<l} \mu_a \omega_a
    + \sum_{a=l}^{a=k} \mu_a \omega_a^2.
  \end{split} 
\end{equation}
From the above equations, Eq.(\ref{kappasigmarelation}) of Section \ref{toymodel} readily follow.

For $k = 2$ everything can be parameterized by two real numbers $\theta_{\pm}$ via $\omega_{\pm} = e^{\pm e^{\theta_{\pm}}}$ (we use $\pm$ as two values of index $a$ in $\omega_a$). For the multiplicity we find $\mu_{\pm} = (1-\omega_{\mp}) / (\omega_{\pm}-\omega_{\mp})$. We list below $\kappa$ and $\Sigma$ for different ranges of $\xi$. For $\xi \le 1/\omega_+$ we have
\begin{equation}
  \kappa = \xi, \quad
  \Sigma = \frac{1}{\xi},
  \label{lowxi}
\end{equation}
for $1/\omega_+ \le \xi \le 1/\omega_-$ we have
\begin{equation}
  \kappa = \xi\frac{(\omega_+-1)\omega_-}{\omega_+-\omega_-} 
  + \frac{1-\omega_-}{\omega_+-\omega_-}, \quad
  \Sigma = \frac{1}{\xi}\frac{(\omega_+-1)\omega_-}{\omega_+-\omega_-} 
  + \frac{\omega_+^2 (1-\omega_-)}{\omega_+-\omega_-},
  \label{middlexi}
\end{equation}
and for $1/\omega_- \le \xi$ we have
\begin{equation}
  \kappa = 1, \quad
  \Sigma =  \omega_+(1-\omega_-) + \omega_-.
  \label{highxi}
\end{equation}
The last expression represents the original gradient variance $\Sigma_0$. Fitting $2$-weight model to empirical data from Section \ref{empirical}, we find $\theta_- = 0.086$ and $\theta_+ = 2.08$. The predicted $\rho(\xi)$ and $\kappa(\xi)$ are described by Eq.(\ref{middlexi}), as the range of $c$ in Fig.(\ref{rhokappavscn}) falls within the middle range $1/\omega_+ < \xi <1/\omega_-$, hence $\alpha = 1$ and $\beta = -1$. These exponents describe the sparsity-variance tradeoff qualitatively correctly, but they lack the quantitative accuracy of power-law fits of Section \ref{empirical} with $\alpha = 0.78$ and $\beta = -1.50$.

\section{Attention spread in transformers: complete account}
\label{attnspreadmodels}

\begin{figure}[h]
  \centering
  \begin{minipage}{0.49\textwidth}
    \begin{minipage}{\textwidth}
      \centering
      \includegraphics[width=\textwidth]{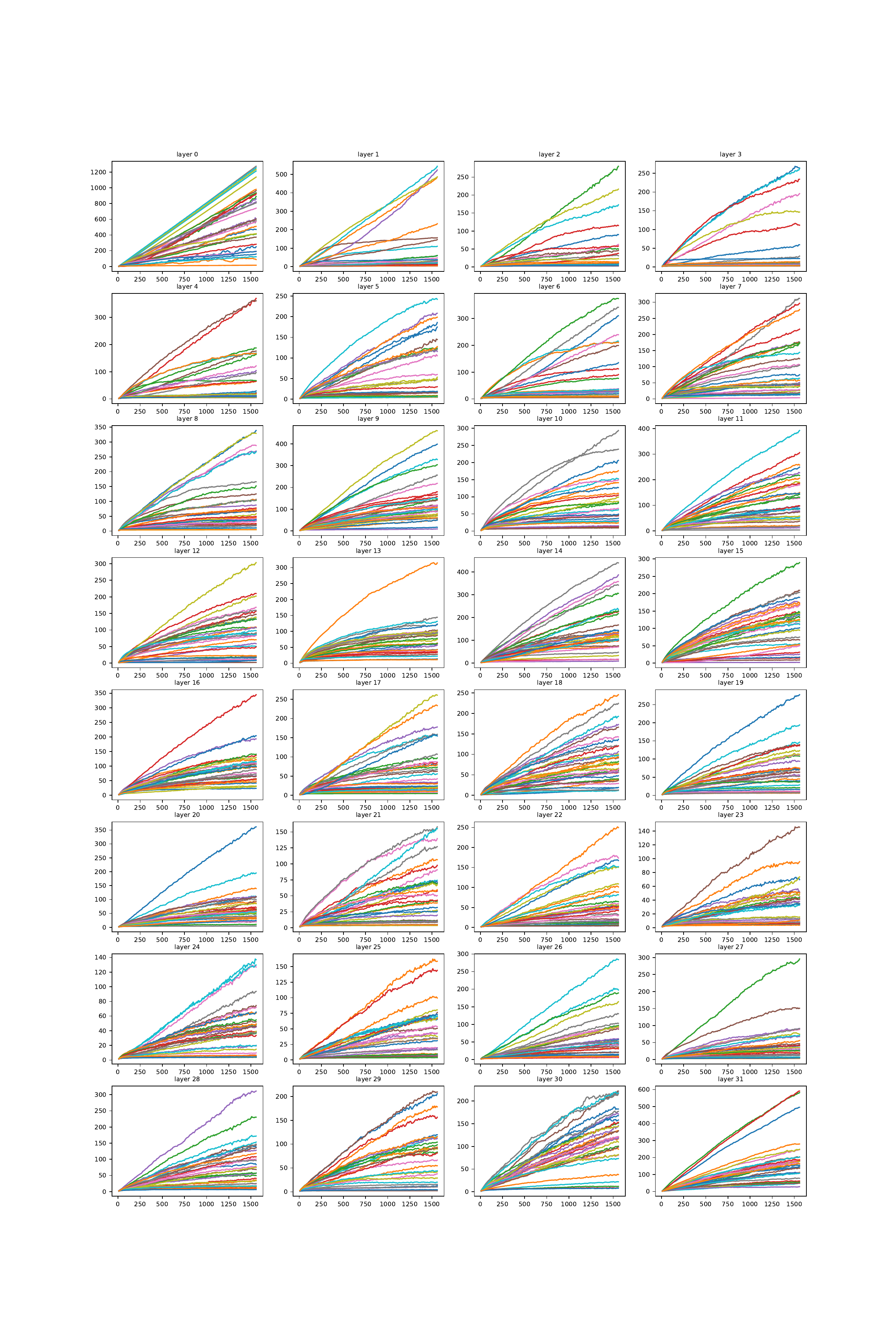}
      \caption{Mistral-7B-v0.1: attention spread $s_i$ vs token position $i$ for every layer and head; 32 layers, 32 attention heads in each layer.}
      \label{attnspreadmistral}
    \end{minipage}

    \begin{minipage}{\textwidth}
      \centering
      \includegraphics[width=\textwidth]{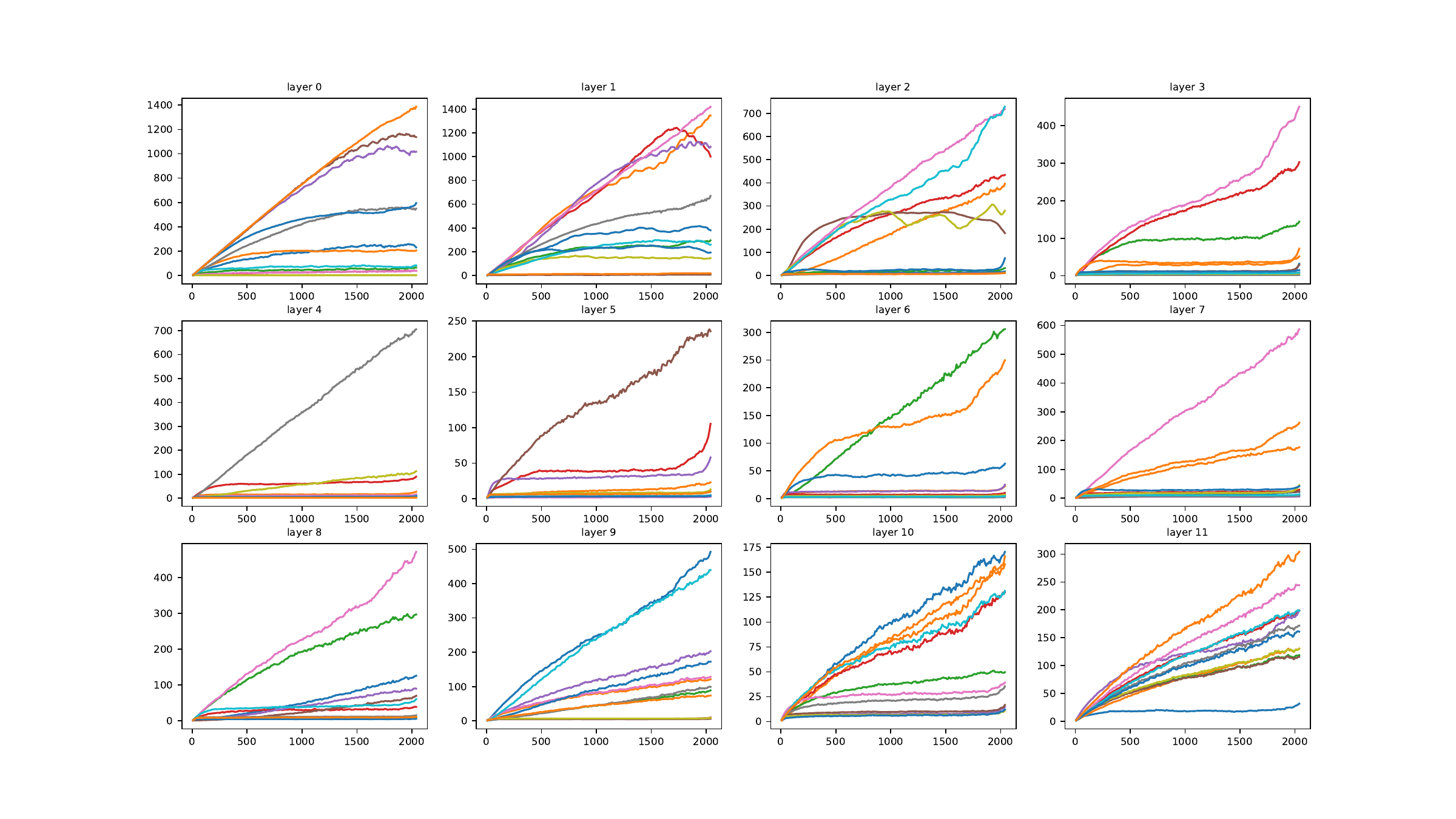}
      \caption{opt-125m: attention spread $s_i$ vs token position $i$ for every layer and head; 12 layers, 12 heads in each layer.}
      \label{attnspreadopt125m}
    \end{minipage}
  \end{minipage}
  \hfill
  \begin{minipage}{0.49\textwidth}
    \begin{minipage}{\textwidth}
      \centering
      \includegraphics[width=\textwidth]{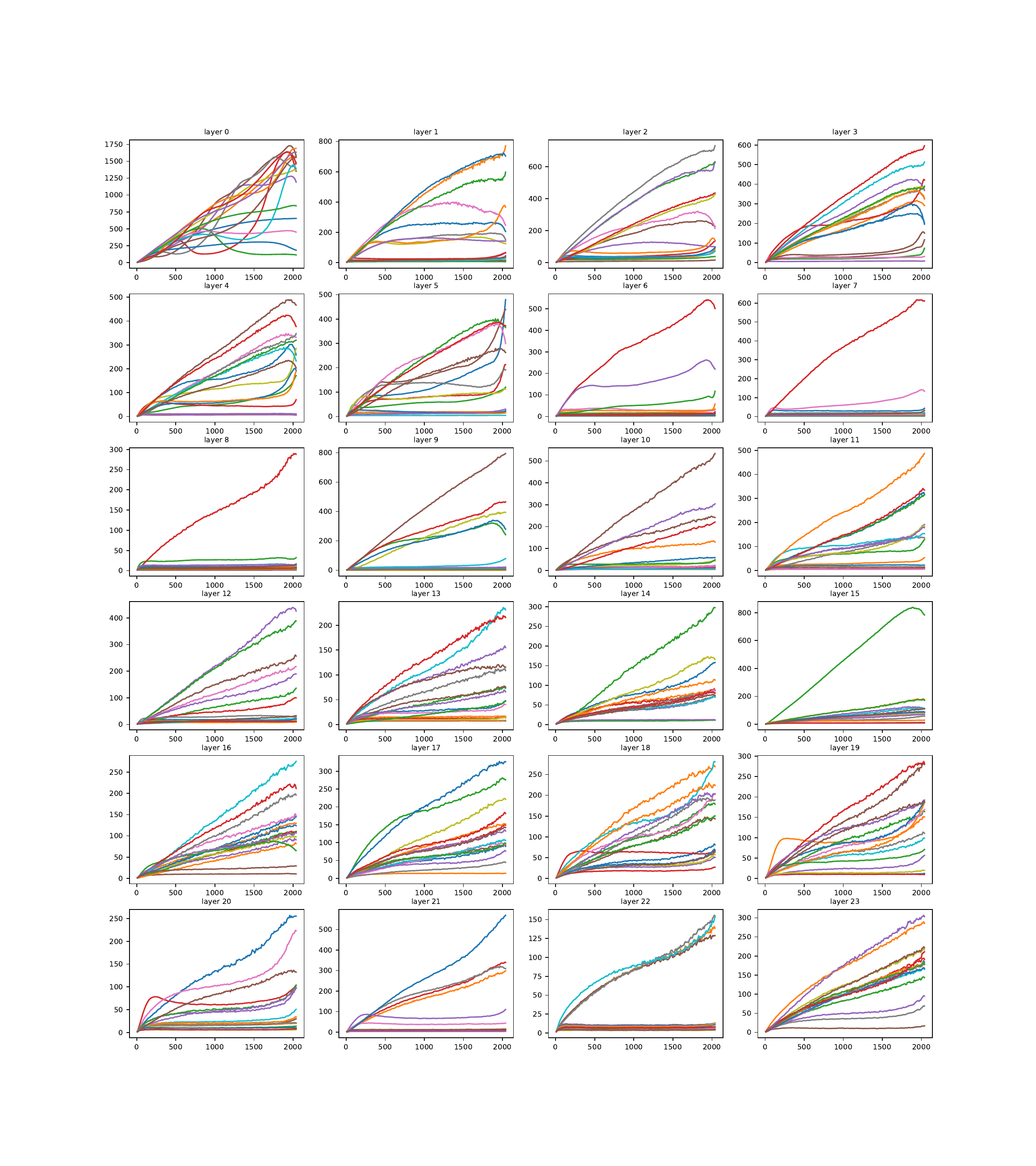}
      \caption{opt-350m: attention spread $s_i$ vs token position $i$ for every layer and head; 24 layers, 16 heads in each layer.}
      \label{attnspreadopt350m}
    \end{minipage}

    \begin{minipage}{\textwidth}
      \centering
      \includegraphics[width=\textwidth]{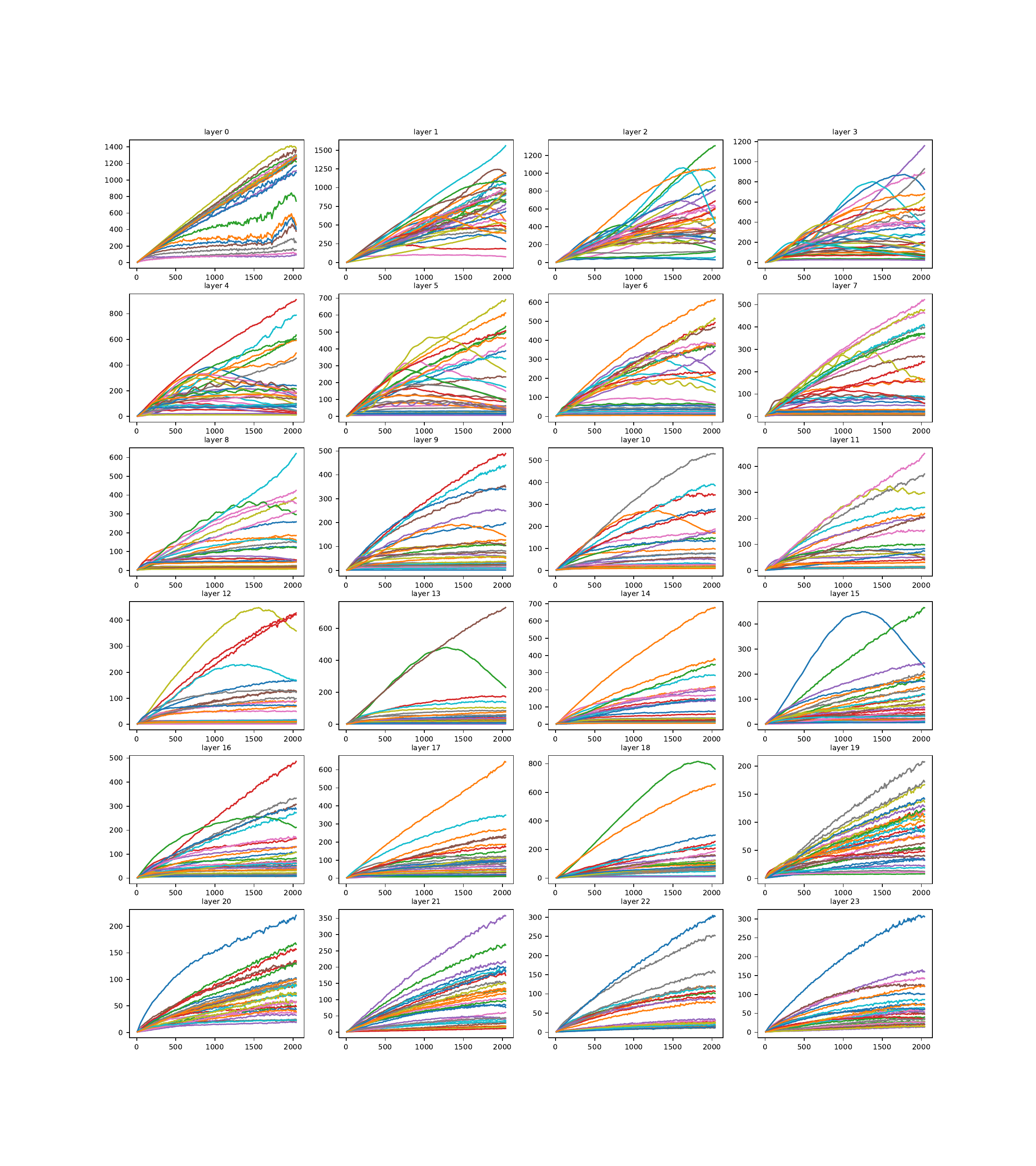}
      \caption{opt-1.3b: attention spread $s_i$ vs token position $i$ for every layer and head; 24 layers, 16 heads in each layer.}
      \label{attnspreadopt1_3b}
    \end{minipage}
  \end{minipage}
\end{figure}

This appendix contains the figures showing the attention spread $s_i$ vs token position $i$ for each attention head, grouped by layers, for all four model that we have considered: {\it opt-125m} in Fig.(\ref{attnspreadopt125m}), {\it opt-350m} in Fig.(\ref{attnspreadopt350m}), {\it opt-1.3b} in Fig.(\ref{attnspreadopt1_3b}), and {\it Mistral-7B-v0.1} in Fig.(\ref{attnspreadmistral}). In all cases we used ${\mathsf p} = 0.9$. The spread $s_i$ data points are plotted in 10 token position increments, averaged over 10 tokens, correspondingly.

\end{document}